\documentclass[runningheads]{llncs}
\usepackage[T1]{fontenc}
\usepackage{graphicx}
\usepackage{hhline}
\usepackage{subfigure}
\usepackage{amsmath,amssymb,amsfonts}
\usepackage{multirow}

\begin{document}
\title{An Explainable Hybrid AI Framework for Enhanced Tuberculosis and Symptom Detection}
%
%
\author{Neel Patel\inst{1} \and
Alexander Wong\inst{2} \and
Ashkan Ebadi\inst{2,3}\orcidID{0000-0002-4542-9105}}
\authorrunning{N. Patel et al.}
%
\institute{Mechanical and Mechatronics Engineering, University of Waterloo, Waterloo ON N2L 3G1, Canada  \\
\and
Systems Design Engineering, University of Waterloo, Waterloo ON N2L 3G1, Canada \\
\and
Digital Technologies, National Research Council Canada, Toronto ON M5T 3J1, Canada \\
\email{ashkan.ebadi@nrc-cnrc.gc.ca}}
\maketitle              
\begin{abstract}
Tuberculosis remains a critical global health issue, particularly in resource-limited and remote areas. Early detection is vital for treatment, yet the lack of skilled radiologists underscores the need for artificial intelligence (AI)-driven screening tools. Developing reliable AI models is challenging due to the necessity for large, high-quality datasets, which are costly to obtain. To tackle this, we propose a teacher--student framework which enhances both disease and symptom detection on chest X-rays by integrating two supervised heads and a self-supervised head. Our model achieves an accuracy of 98.85\% for distinguishing between COVID-19, tuberculosis, and normal cases, and a macro-F1 score of 90.09\% for multilabel symptom detection, significantly outperforming baselines. The explainability assessments also show the model bases its predictions on relevant anatomical features, demonstrating promise for deployment in clinical screening and triage settings.

\end{abstract}
\section{Introduction}
Tuberculosis (TB) continues to be a major global health challenge, despite being both preventable and curable \cite{patel2024}. Transmitted via aerosols of transmission of mycobacterium tuberculosis, TB remains particularly burdensome in low– and middle–income regions, ranking second only to COVID‑19 among infectious causes of death in 2022 \cite{who2023,who2024}. The risk of developing active TB is highest within the first two years post-infection--approximately 5\% progress during this period, with an additional \(\sim\)5\% later in life--highlighting the critical need for timely diagnosis and treatment \cite{un2022,cdc2025}. Untreated pulmonary TB has a high fatality rate (pre‑chemotherapy estimates around 70\% for smear‑positive cases), whereas current 4–6 month treatment regimens achieve a global treatment success rate of \(\sim\)88\% for drug‑susceptible TB in 2022 \cite{who2024treat}. Despite reaching record notifications of 7.5 million in 2022 and 8.2 million in 2023, global efforts still fall short of the End TB Strategy milestones \cite{who2015endtb,who2023cases}.

Chest radiography (CXR) is the workhorse for TB screening  \cite{diaz2020} due to its rapidity and widespread availability, yet the quality of interpretation varies with reader expertise and workload, particularly in high‑burden settings \cite{chexaid2020}. Reflecting this reality, recent World Health Organization (WHO) guidance supports the use of computer‑aided detection (CAD) to interpret digital CXR for TB screening and triage in people aged 15 and older  \cite{who2021screen,who2025cad}.

Identifying and screening populations at risk is crucial for effectively controlling the spread of the TB disease \cite{wong2022}. To bridge the data and trust gaps that limit robust CAD implementation, we have developed an explainable, semi‑supervised framework for CXR that performs both disease triage and radiographic symptom detection within a unified model. Our approach is based on a ``distillation for self-supervision'' paradigm (DISTL) \cite{park2022}, employing a ViT‑Small teacher--student backbone optimized with a DINO \cite{dino2021} self‑supervised head, alongside two supervised heads for learning disease and symptoms.  We leverage CheXpert‑pretrained \cite{chexpert2019} CXR weights for domain‑specific initialization and utilize multi‑crop training to expose the network to both global lung context and fine local patterns. The model’s decision-making process is further validated by comparing Gradient-weighted Class Activation Mapping (Grad‑CAM) \cite{gradcam2017} saliency maps to dataset bounding boxes for symptom findings, ensuring alignment with plausible thoracic anatomy. 

Our experiments on a comprehensive dataset, including TB, COVID‑19, normal controls, and seven symptom labels, demonstrate that the proposed model outperforms strong CNN baselines across both tasks. It achieves near‑ceiling performance in disease classification and substantial gains in symptom detection, particularly for small or subtle findings, with Grad‑CAM overlays focusing within annotated lung and pleural regions. These results indicate that integrating self‑supervised distillation with joint disease–symptom supervision produces features that are both discriminative and clinically relevant, supporting deployment in real‑world TB screening workflows where sensitivity, specificity, and explainability are crucial.

\section{Data}
A composite chest-radiograph corpus was assembled from four open-access sources to cover disease, healthy control, and finding-level tasks. Tuberculosis-positive images ($n=2,141$) were collected from the Montgomery ($n=58$) \cite{montgomery2014}, Shenzhen ($n=336$) \cite{montgomery2014}, Belarus TB-Portals ($n=1,047$) \cite{belarus} and Rahman et al. ($n=700$) \cite {rahman2020} cohorts. Normal controls ($n=3,500$) were taken from the Rahman collection \cite {rahman2020}. Additionally, $4,000$ positive COVID-19 images were collected from the COVIDx-CXR4 repository \cite{covidx-cxr}. Finally, $15,000$ images bearing at least one of seven radiological findings—infiltration, effusion, atelectasis, nodule, mass, pneumothorax, and consolidation—were sampled from the NIH ChestXray14 archive \cite{nihchest} (see Table~\ref{tab:dataset_distribution}). 

\begin{table}[htbp]
\caption{Data distribution -- Diseases and Symptoms.}
\centering
\begin{tabular}{|c|c|c|c|c|c|}
\hline
\textbf{No} & \textbf{Source} & \textbf{Normal} & \textbf{TB} & \textbf{COVID} & \textbf{Symptoms} \\
\hline
\textbf{1} & Montgomery \cite{montgomery2014} & 0 & 58 & 0 & -- \\
\hline
\textbf{2} & Shenzhen \cite{montgomery2014} & 0 & 336 & 0 & -- \\
\hline
\textbf{3} & Belarus \cite{belarus} & 0 & 1,047 & 0 & -- \\
\hline
\textbf{4} & Rahman et al. \cite{rahman2020} & 3,500 & 700 & 0 & -- \\
\hline
\textbf{5} & COVIDx-CXR \cite{covidx-cxr} & 0 & 0 & 4,000 & -- \\
\hline
\textbf{6} & NIH ChestXray \cite{nihchest} & -- & -- & -- & 15,000 \\
\hhline{|=|=|=|=|=|=|}
 \multicolumn{2}{|c|}{\textbf{Total}} & 3,500 & 2,141 & 4,000 & 15,000 \\
\hline
 \multicolumn{2}{|c|}{\textbf{After quality control}} & 3,163 & 1,904 & 2,358 & 12,024 \\
\hline
\end{tabular}
\label{tab:dataset_distribution}
\end{table}

Table~\ref{tab:symptoms} presents the distribution of symptoms within the dataset, along with their total counts and the percentage of images in which they appear. The total percentage exceeds 100\% since the dataset allows for multi-label classification, meaning an image can exhibit more than one symptom simultaneously.

\begin{table}[htbp]
\caption{Symptoms distribution.}
\centering
\begin{tabular}{|l|c|c|}
\hline
\textbf{Symptom} & \textbf{Total Count} & \textbf{\% of Images} \\
\hline
Infiltration & 4,379 & 36.4\% \\
Effusion & 3,351 & 27.9\% \\
Atelectasis & 3,311 & 27.5\% \\
Nodule & 2,105 & 17.5\% \\
Mass & 1,881 & 15.6\% \\
Pneumothorax & 1,740 & 14.5\% \\
Consolidation & 1,465 & 12.2\% \\
\hline
\textbf{Total} & \textbf{18,232} & \textbf{151.6\%}$^\dagger$ \\
\hline
\multicolumn{3}{l}{$^{\dagger}$Sum exceeds 100\% due to multi-label nature.}\\
\end{tabular}
\label{tab:symptoms}
\end{table}

Table~\ref{tab:multi_symptoms} provides information on the distribution of images based on the number of symptoms they exhibit, with a total of 12,024 images. As seen, the majority of images (62.6\%) contain a single symptom, while fewer images contain multiple symptoms, ranging from two symptoms (26.3\%) to seven symptoms (0.01\%).

\begin{table}[htbp]
\caption{Multi-symptoms distribution.}
\centering
\begin{tabular}{|l|c|c|}
\hline
 & \textbf{Count} & \textbf{Percentage} \\
\hline
\textbf{Total Images} & 12,024 & 100.0\% \\
\hline
\multicolumn{3}{|c|}{\textbf{Multi-Symptom Distribution}} \\
\hline
Single symptom & 7,530 & 62.6\% \\
Two symptoms & 3,157 & 26.3\% \\
Three symptoms & 1,030 & 8.6\% \\
Four symptoms & 251 & 2.1\% \\
Five symptoms & 43 & 0.4\% \\
Six symptoms & 12 & 0.1\% \\
Seven symptoms & 1 & 0.01\% \\
\hline
\end{tabular}
\label{tab:multi_symptoms}
\end{table}

\section{Methods}
\begin{figure}[http]
\centerline{\includegraphics[width=0.8\textwidth]{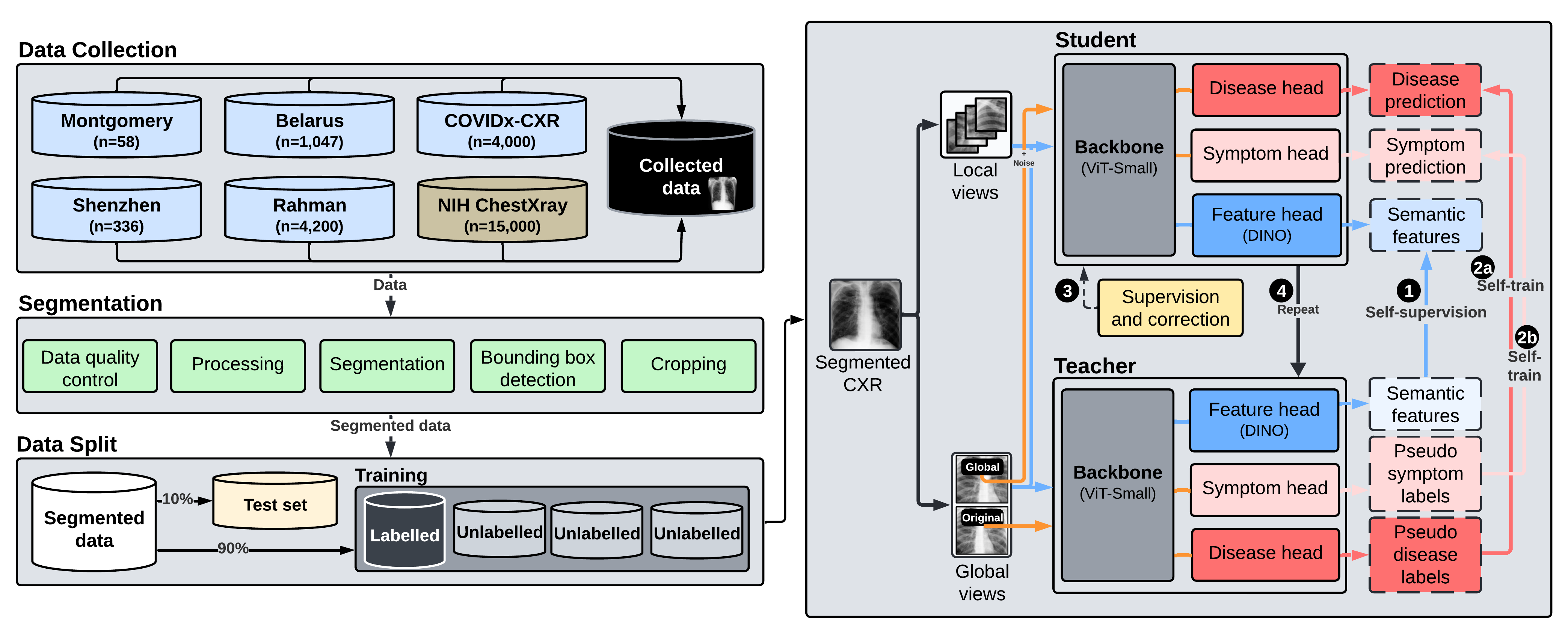}}
\caption{High-level conceptual flow of the analysis.}
\label{fig:concep_flow}
\end{figure}

Fig.~\ref{fig:concep_flow} presents a high-level conceptual flow of the data analysis process, beginning with data collection and integration, followed by segmentation, and concluding with the model training phase. The components depicted in the figure are described in detail in this section, highlighting the methodologies and techniques employed at each stage.

\subsection{Lung Segmentation}
We employ U-Net \cite{unet2015}, a symmetric encoder–decoder convolutional network with skip connections that preserves fine edge details while maintaining global context, making it the standard choice for biomedical image segmentation. At the start of the process, we load a pretrained checkpoint \cite{ovcharenko} and keep it in evaluation mode throughout the pipeline. Each radiograph is converted to an 8-bit greyscale format and resized to 225$\times$225 pixels before being processed by the network; the output per-pixel logits are then transformed into a binary mask using arg-max and nearest-neighbour up-sampling.

\begin{figure}[htbp]
    \centering
    \subfigure[Normal case, original]
    {
        \includegraphics[width=1.in]{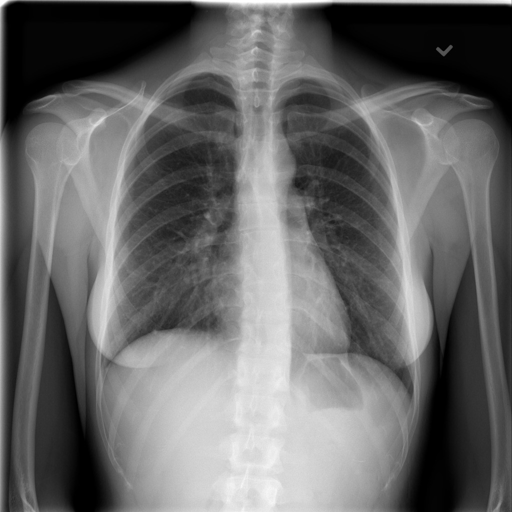}
        \label{fig3a}
    }
    \subfigure[Normal case, cropped]
    {
        \includegraphics[width=1.in]{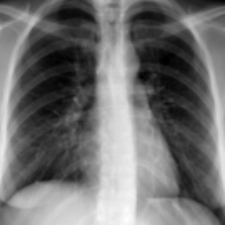}
        \label{fig3b}
    }
    
    \subfigure[Tuberculosis case, original]
    {
        \includegraphics[width=1.in]{"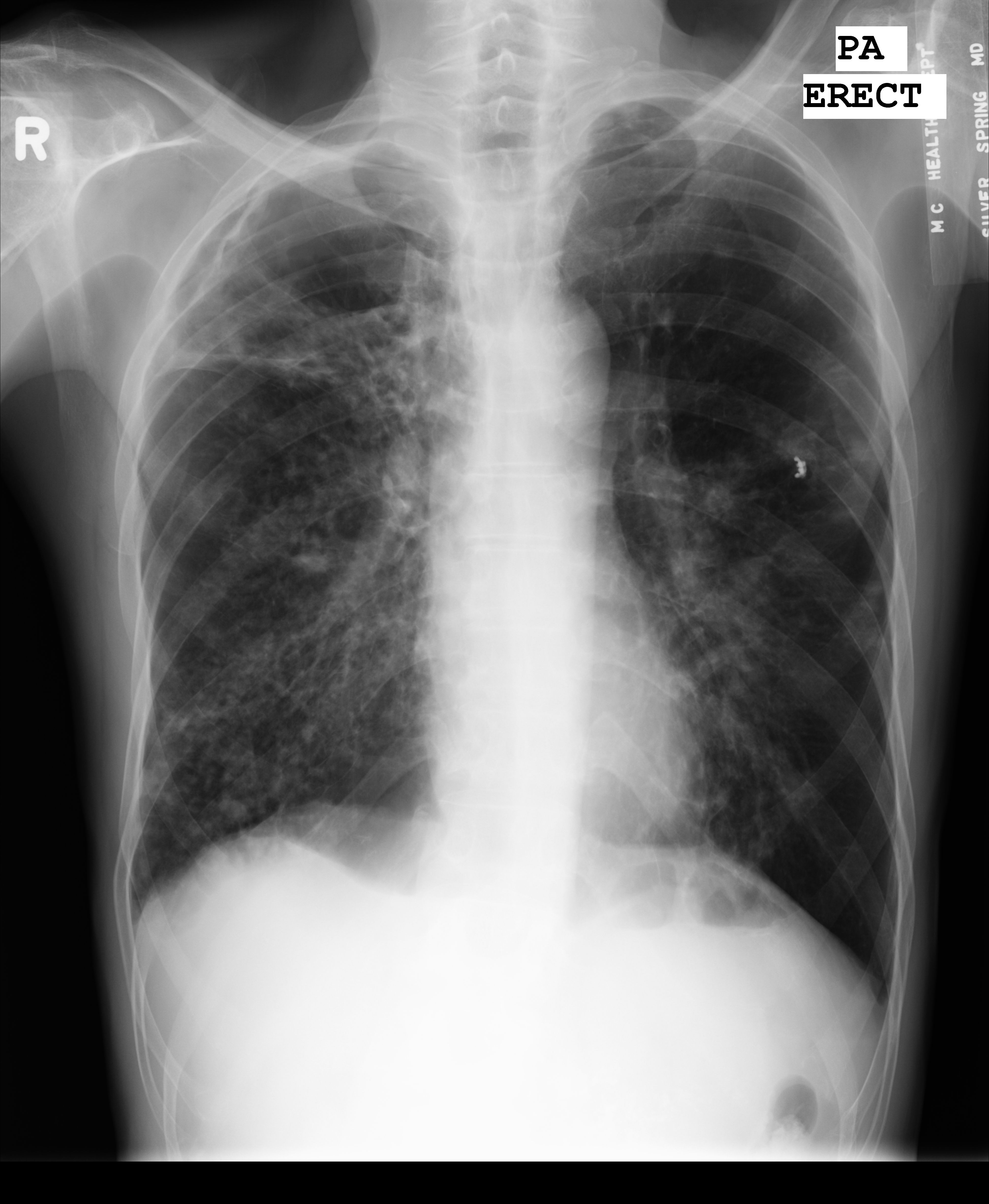"}
        \label{fig3a}
    }
    \subfigure[Tuberculosis case, cropped]
    {
        \includegraphics[width=1.in]{"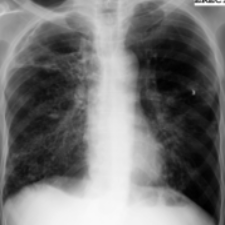"}
        \label{fig3b}
    }
    \caption{Sample images in the dataset.}
    \label{fig_sample_images}
\end{figure}

We conduct rigorous quality control to ensure the integrity of our segmentation masks. Masks are retained only if (i) lung pixels occupy between 8\% and 90\%of the image, and (ii) there are at least two external contours present. Accepted masks define the tightest bounding box around the two largest contours. The corresponding region is cropped from the original CXR to focus subsequent analysis on parenchymal tissue while minimizing background artifacts. Any image failing these automatic checks or showing residual issues during subsequent manual review (such as embedded lead markers) is discarded. This procedure resulted in the elimination of 337 normal, 237 tuberculosis, 1,642 COVID-19, and 2,976 symptom-labelled images, leaving a curated dataset of 1,904 tuberculosis, 3,163 normal, 2,358 COVID-19, and 12,024 symptom CXRs, totalling 19,449 images. Fig.~\ref{fig_sample_images} presents sample images from the dataset, showcasing both original and cropped versions, demonstrating how the segmentation and cropping process isolates relevant regions while reducing background noise.

\subsection{Data Split}
The curated dataset was split on a patient-wise basis: 10\% was reserved as an independent unseen test set, and the remaining 90\% constituted the training pool. Within this training pool, 30\% was treated as labelled data, and the rest was evenly divided into three unlabeled folds for self-supervised distillation.

\subsection{Network Architecture}
We adopt the DISTL framework \cite{park2022}, configuring a teacher--student network architecture using a ViT-Small backbone for both teacher and student models. 
Images are transformed into $384$-dimensional tokens via a strided convolution with a patch size of $8$, and processed through $12$ transformer blocks, featuring $6$ heads, LayerNorm ($\varepsilon=10^{-6}$), and a stochastic depth $0.1$ applied solely to the student network. Each backbone operates within a multi-crop pipeline, 
where the student receives two global crops (the original and a $256$-pixel crop with a scale range of $(0.75,\,1.0)$) along with multiple local crops (defaulting to $8$ at $128$ pixels with a scale range of$(0.2,\,0.6)$), while the teacher processes only the two global crops.

Three distinct heads are attached to the backbone: (1) a DINO \cite{dino2021} projection head, consisting of a 3-layer multilayer perception (MLP) with GELU activation, featuring a \texttt{hidden\_dim} of $2048$, \texttt{bottleneck\_dim} of $256$, and a weight-normalized output layer with \texttt{out\_dim} of $65{,}536$, which is used for the self-supervised learning objective; (2) a disease classificiation head, a 3-layer MLP with ReLU activation as: $384\!\rightarrow\!256\!\rightarrow\!256\!\rightarrow\!3$, designed to produce logits for a three-class disease classifier; and (3) a parallel symptom classification head, another 3-layer MLP with ReLU activation configured as $384\!\rightarrow\!256\!\rightarrow\!256\!\rightarrow\!7$, which outputs logits for a multi-label set of radiological symptoms, i.e., infiltration, effusion, atelectasis, nodule, mass, pneumothorax, and consolidation. Both networks are initialized with ViT-Small weights pre-trained on the CheXpert dataset \cite{chexpert2019}.

\subsection{Training Strategy}
Our training strategy integrates self-supervision and self-training techniques. The DINO framework is used to align student and teacher features across different views, while knowledge distillation is employed to transfer the teacher's predictions to the student. This is achieved using temperature-scaled KL divergence (with a temperature $\tau=2.0$) for the disease classification head, and focal binary cross-entropy (BCE) loss (with a gamma value of $\gamma=2.0$) applied to sigmoid-activated teacher logits for the symptom classification head. During the initial training epochs ($<\texttt{ssl\_epoch}$), these losses are combined with a mixing coefficient $\lambda$ (default set to $0.5$), gradually transitioning to pure distillation without the DINO loss. The teacher model's parameters are updated using an exponential moving average (EMA) of the student's parameters, following a cosine momentum schedule ranging from $0.9995$ to $1.0$. Additionally, every $500$ iterations, a labeled ``correction'' phase is introduced, applying cross-entropy loss to the disease head and focal BCE to the symptom head. This phase incorporates per-label \texttt{pos\_weight} based on prevalence, clipped at $20.0$, and employs a weighted sampler that increases the sampling frequency (by 3 times) of images containing at least one positive symptom. This comprehensive approach allows the model to effectively learn both disease states and radiological symptoms within a unified teacher--student pipeline.

\subsection{Network Training Process}
The network training process is structured to extract features and perform multi-disease classification and radiological symptom detection.  We leverage the DINO head for feature extraction, the CLS head for multi-disease classification, and a symptom head, tailored to detect seven specific radiological findings. 

\subsubsection{Phase 1: Supervised Pretraining}
In the initial phase, we focus on developing robust baseline representations using solely labeled data. A three-tiered data augmentation strategy is employed to enhance robustness under varied imaging conditions. The first global crop undergoes minimal transformation, primarily normalization using ImageNet \cite{imagenet} statistics.
The second global crop receives moderate augmentation, including random resized cropping ($256{\times}256$ pixels, scale $0.75$--$1.0$), horizontal flipping ($p{=}0.5$), rotation ($\pm 15^{\circ}$), auto-contrast, equalization, and Gaussian blur (each with $p{=}0.3$). Local crops ($8$ crops of $128{\times}128$ pixels, scale $0.2$--$0.6$) undergo stronger augmentation with higher probabilities ($p{=}0.5$) to capture fine-grained pathological details crucial for symptom detection. The student minimizes a weighted combination of losses: cross-entropy with inverse-frequency class weights for disease classification and focal binary cross-entropy (Focal BCE, $\gamma=2.0$) for symptom detection. The combined loss is formulated as:
\[
\mathcal{L} \;=\; 0.25\,\mathcal{L}_{\text{disease}} \;+\; 0.75\,\mathcal{L}_{\text{symptom}},
\]
prioritizing symptom learning. Symptom detection incorporates per-label positive weights derived from class prevalence, clipped at $50.0$, with a $1.5{\times}$ additional boost for rare symptoms, such as pneumothorax and consolidation, capped at $75.0$. Optimization is performed using AdamW with cosine annealing, starting from a learning rate of $5{\times}10^{-5}$ to $10^{-6}$, and a fixed weight decay of $0.01$.

\subsubsection{Phase 2: Semi-Supervised DISTL Training}
In this phase, both teacher and student models are initialized from Phase~1. The DISTL framework is fully utilized, leveraging both labeled and unlabeled data through three successive training runs (folds), each incorporating progressively larger portions of unlabeled data. We use two primary loss mechanisms:

\begin{enumerate}
    \item \textbf{DINOLoss:} Aligns feature representations between teacher and student across views. The teacher processes only global views, while the student processes both global and local views. The loss is computed using cross-entropy between softmax outputs after centring and sharpening with temperature scheduling. 
    \item \textbf{Knowledge distillation:} Transfers the knowledge from the teacher to the student via two paths: KL divergence with temperature scaling ($\tau=2.0$) for disease classification, and Focal BCE for symptom detection, comparing student logits against sigmoid-activated teacher predictions. The distillation loss is given by:
\[
\mathcal{L}_{\text{distill}} \;=\; \tfrac{1}{2}\,\Big(\mathcal{L}_{\text{KL}} \;+\; \mathcal{L}_{\text{symptom}}\Big).
\]
During early epochs ($<\texttt{ssl\_epoch}$), the total loss is:
\[
\mathcal{L} \;=\; \lambda\,\mathcal{L}_{\text{DINO}} \;+\; (1-\lambda)\,\mathcal{L}_{\text{distill}}, \qquad \lambda=0.5,
\]
after which training transitions to pure distillation ($\mathcal{L}=\mathcal{L}_{\text{distill}}$). Student parameters are updated by backpropagation, while teacher parameters are updated using EMA with a cosine momentum schedule ($0.9995$ to $1.0$). Across folds, the student is warm-started from its latest checkpoint, and the teacher from its EMA weights, stabilizing training as the proportion of unlabeled data increases.
\end{enumerate}

\subsubsection{Phase 3: Correction Phase}
Every $500$ iterations, a supervised correction phase is implemented to counteract pseudo-label noise and maintain model accuracy. During this phase, the student is trained with ground-truth labels using cross-entropy for disease classification and Focal BCE for symptom detection, with positive weights clipped at $20.0$. To address class imbalance, a weighted sampler increases the selection probability of images containing at least one positive symptom by threefold.

\subsection{Performance Evaluation}
For performance evaluation, disease classification is assessed using precision, recall, F1-score, and accuracy metrics across three categories: COVID-19, Normal, and Tuberculosis. Symptom detection is evaluated in a multi-label context, employing per-symptom and average precision, recall, and F1-scores. We benchmark our model against four convolutional neural network (CNN) baselines-- vanilla CNN, VGG16, ResNet-18, ResNet-50-- all of which are fine-tuned following the same multi-class classification protocol.

\section{Results}\label{sec:results}
In this section, we present results from the held‑out test split for two key tasks: (i) three‑class disease classification, and (ii) multilabel detection of seven radiological symptoms. Metrics are calculated per class/label on the test set, unless otherwise specified, and we also provide macro‑averages--unweighted means across classes/labels-- to facilitate comparability.

\subsection{Three‑class Disease Classification}
Table~\ref{tab:disease_cls} showcases a comparison between our teacher--student ViT model and four CNN baselines, all trained and evaluated on the same dataset, with the highest values highlighted in bold. Among the baselines, VGG16 emerges as the strongest, achieving an overall accuracy of 96.84\% and macro‑F1 score of 96.86\%. In contrast, our model significantly outperforms these benchmarks with an accuracy of 98.85\% and a macro‑F1 score of 98.89\%, marking improvements of +2.01 and +2.03 percentage points over the best baseline, respectively. When examining class‑wise performance, our model delivers F1 scores of 98.94\% for normal cases, 99.44\% for tuberculosis, and 98.30\% for COVID‑19, surpassing the best baseline in each class by +2.08, +2.26, and +1.75 percentage points, respectively. Additionally, macro precision and recall metrics are enhanced, with our model achieving 98.94\% and 98.85\%, compared to the best baseline's 96.99\% and 96.75\%. In terms of recall, our model demonstrates marked improvements across all three classes: Normal sees an increase of +1.05 (98.94\% vs. 97.89\%), TB rises by +0.57 (98.88\% vs. 98.31\%), and COVID‑19 improves by +1.28 (98.72\% vs. 97.44\%). We attribute these improvements to: (i) a lung‑focused preprocessing approach that reduces non‑parenchymal artifacts, and (ii) the integration of multi‑crop self‑supervision and teacher--student distillation, which enhances the stability of representation learning from limited labels.

\begin{table}[htbp]
\caption{Model performance comparison on 3-class disease classification.}
\centering
\begin{tabular}{|l|l|c|c|c|c|}
\hline
\textbf{Model} & \textbf{Class} & \textbf{Precision} & \textbf{Recall} & \textbf{F1} & \textbf{Accuracy} \\
\hline
\multirow{3}{*}{\textbf{Vanilla CNN}} 
 & Normal    & 73.41\% & 93.31\% & 82.17\% & \multirow{3}{*}{80.60\%} \\
\cline{2-5}
 & TB        & 91.84\% & 75.84\% & 83.08\% &  \\
\cline{2-5}
 & COVID  & 85.64\% & 68.80\% & 76.30\% &  \\
\hline
\multirow{3}{*}{\textbf{VGG16}} 
 & Normal    & 95.86\% & 97.89\% & 96.86\% & \multirow{3}{*}{96.84\%} \\
\cline{2-5}
 & TB        & 97.73\% & 96.63\% & 97.18\% &  \\
\cline{2-5}
 & COVID  & 97.39\% & 95.73\% & 96.55\% &  \\
\hline
\multirow{3}{*}{\textbf{ResNet18}} 
 & Normal    & 95.02\% & 94.01\% & 94.51\% & \multirow{3}{*}{92.10\%} \\
\cline{2-5}
 & TB        & 96.75\% & 83.71\% & 89.76\% &  \\
\cline{2-5}
 & COVID  & 86.21\% & 96.15\% & 90.91\% &  \\
\hline
\multirow{3}{*}{\textbf{ResNet50}} 
 & Normal    & 98.86\% & 91.55\% & 95.06\% & \multirow{3}{*}{95.26\%} \\
\cline{2-5}
 & TB        & 90.67\% & 98.31\% & 94.34\% &  \\
\cline{2-5}
 & COVID  & 95.00\% & 97.44\% & 96.20\% &  \\
\hline
\multirow{3}{*}{\textbf{Our model}} 
 & Normal    & \textbf{98.94\%} & \textbf{98.94\%} & \textbf{98.94\%} & \multirow{3}{*}{\textbf{98.85\%}} \\
\cline{2-5}
 & TB        & \textbf{100.00\%} & \textbf{98.88\%} & \textbf{99.44\%} &  \\
\cline{2-5}
 & COVID  & \textbf{97.88\%} & \textbf{98.72\%} & \textbf{98.30\%} &  \\
\hline
\end{tabular}
\label{tab:disease_cls}
\end{table}

\subsection{Symptom‑wise multilabel detection}
Table~\ref{tab:sym} presents a detailed summary of per‑symptom precision, recall, and F1 scores for the seven radiological findings, with the highest values highlighted in bold. Our model demonstrates significant advancements in multi-label detection, achieving a macro‑F1 score of 90.09\%, along with substantial improvements in macro precision (86.90\%) and macro recall (93.60\%), registering absolute gains of +31.76, +30.50, and +28.13 percentage points over the best-performing baseline, respectively. Notably, improvements are consistent across all symptoms, with the most significant F1 score gains seen in detecting smaller or more subtle anomalies: nodule (+40.58; 87.05\% vs.\ 46.47\%), effusion (+37.17; 89.02\% vs.\ 51.85\%), mass (+35.37; 85.08\% vs.\ 49.71\%), consolidation (+28.98; 84.49\% vs.\ 55.51\%), infiltration (+27.03; 93.36\% vs.\ 66.33\%), and atelectasis (+25.49; 94.75\% vs.\ 69.26\%). Pneumothorax also improves strongly (+14.75; 96.91\% vs.\ 82.16\%). Recall gains are particularly marked for nodule (+30.77), effusion (+23.81), mass (+25.56), and infiltration (+14.89). These consistent improvements highlight the efficacy of the proposed approach, underscoring the effectiveness of incorporating multi‑crop local views for detailed pathology detection, class‑balanced sampling with positive re‑weighting for symptom classification, and the DISTL‑style self‑training approach to leverage both labeled and unlabeled data.

\begin{table}[htbp]
\caption{Symptom-wise performance (multilabel detection).}
\centering
\begin{tabular}{|l|l|c|c|c|}
\hline
\textbf{Model} & \textbf{Symptom} & \textbf{Precision} & \textbf{Recall} & \textbf{F1} \\
\hline
\multirow{7}{*}{\textbf{Vanilla CNN}}
 & Pneumothorax   & 67.07\% & 57.89\% & 62.15\% \\
\cline{2-5}
 & Consolidation  & 42.20\% & 81.11\% & 55.51\% \\
\cline{2-5}
 & Nodule         & 30.00\% & 46.15\% & 36.36\% \\
\cline{2-5}
 & Mass           & 14.36\% & 58.89\% & 23.09\% \\
\cline{2-5}
 & Atelectasis    & 60.58\% & 80.84\% & 69.26\% \\
\cline{2-5}
 & Effusion       & 26.67\% & 28.57\% & 27.59\% \\
\cline{2-5}
 & Infiltration   & 44.13\% & 71.76\% & 54.65\% \\
\hline
\multirow{7}{*}{\textbf{VGG16}}
 & Pneumothorax   & 84.44\% & 80.00\% & 82.16\% \\
\cline{2-5}
 & Consolidation  & 58.82\% & 44.44\% & 50.63\% \\
\cline{2-5}
 & Nodule         & 36.23\% & 54.95\% & 43.67\% \\
\cline{2-5}
 & Mass           & 51.81\% & 47.78\% & 49.71\% \\
\cline{2-5}
 & Atelectasis    & 64.07\% & 74.68\% & 68.97\% \\
\cline{2-5}
 & Effusion       & 46.67\% & 58.33\% & 51.85\% \\
\cline{2-5}
 & Infiltration   & 52.75\% & 73.28\% & 61.34\% \\
\hline
\multirow{7}{*}{\textbf{ResNet18}}
 & Pneumothorax   & 46.15\% & 50.53\% & 48.24\% \\
\cline{2-5}
 & Consolidation  & 45.83\% & 61.11\% & 52.38\% \\
\cline{2-5}
 & Nodule         & 23.03\% & 41.76\% & 29.69\% \\
\cline{2-5}
 & Mass           & 53.25\% & 45.56\% & 49.10\% \\
\cline{2-5}
 & Atelectasis    & 52.24\% & 83.12\% & 64.16\% \\
\cline{2-5}
 & Effusion       & 27.80\% & 67.86\% & 39.45\% \\
\cline{2-5}
 & Infiltration   & 59.57\% & 74.81\% & 66.33\% \\
\hline
\multirow{7}{*}{\textbf{ResNet50}}
 & Pneumothorax   & 51.92\% & 56.84\% & 54.27\% \\
\cline{2-5}
 & Consolidation  & 40.82\% & 44.44\% & 42.55\% \\
\cline{2-5}
 & Nodule         & 37.33\% & 61.54\% & 46.47\% \\
\cline{2-5}
 & Mass           & 27.84\% & 60.00\% & 38.03\% \\
\cline{2-5}
 & Atelectasis    & 54.31\% & 87.99\% & 67.16\% \\
\cline{2-5}
 & Effusion       & 29.44\% & 63.10\% & 40.15\% \\
\cline{2-5}
 & Infiltration   & 43.59\% & 84.35\% & 57.48\% \\
\hline
\multirow{7}{*}{\textbf{Our model}}
 & Pneumothorax   & \textbf{94.95\%} & \textbf{98.95\%} & \textbf{96.91\%} \\
\cline{2-5}
 & Consolidation  & \textbf{81.44\%} & \textbf{87.78\%} & \textbf{84.49\%} \\
\cline{2-5}
 & Nodule         & \textbf{82.35\%} & \textbf{92.31\%} & \textbf{87.05\%} \\
\cline{2-5}
 & Mass           & \textbf{84.62\%} & \textbf{85.56\%} & \textbf{85.08\%} \\
\cline{2-5}
 & Atelectasis    & \textbf{90.29\%} & \textbf{99.68\%} & \textbf{94.75\%} \\
\cline{2-5}
 & Effusion       & \textbf{86.52\%} & \textbf{91.67\%} & \textbf{89.02\%} \\
\cline{2-5}
 & Infiltration   & \textbf{88.14\%} & \textbf{99.24\%} & \textbf{93.36\%} \\
\hline
\end{tabular}
\label{tab:sym}
\end{table}


\subsection{Symptom Classification Explainability}
\begin{figure}[htbp]
    \centering

    \subfigure[Mass, original]
    {
        \includegraphics[width=1.1in]{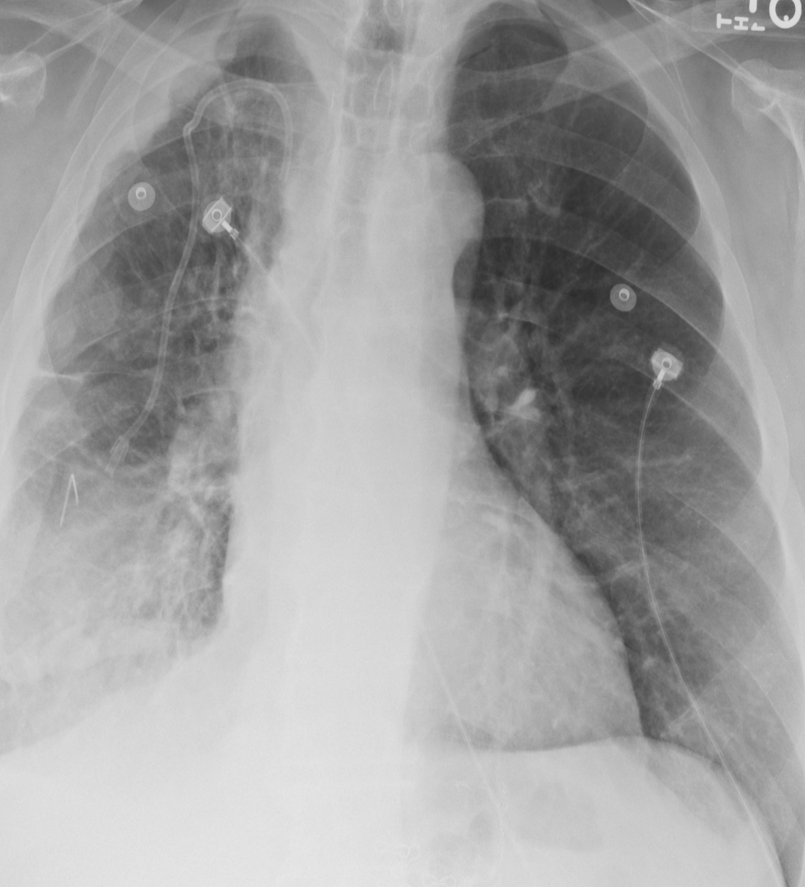}
        \label{fig:mass_orig}
    }
    \subfigure[Mass, Grad-CAM + bbox]
    {
        \includegraphics[width=1.1in]{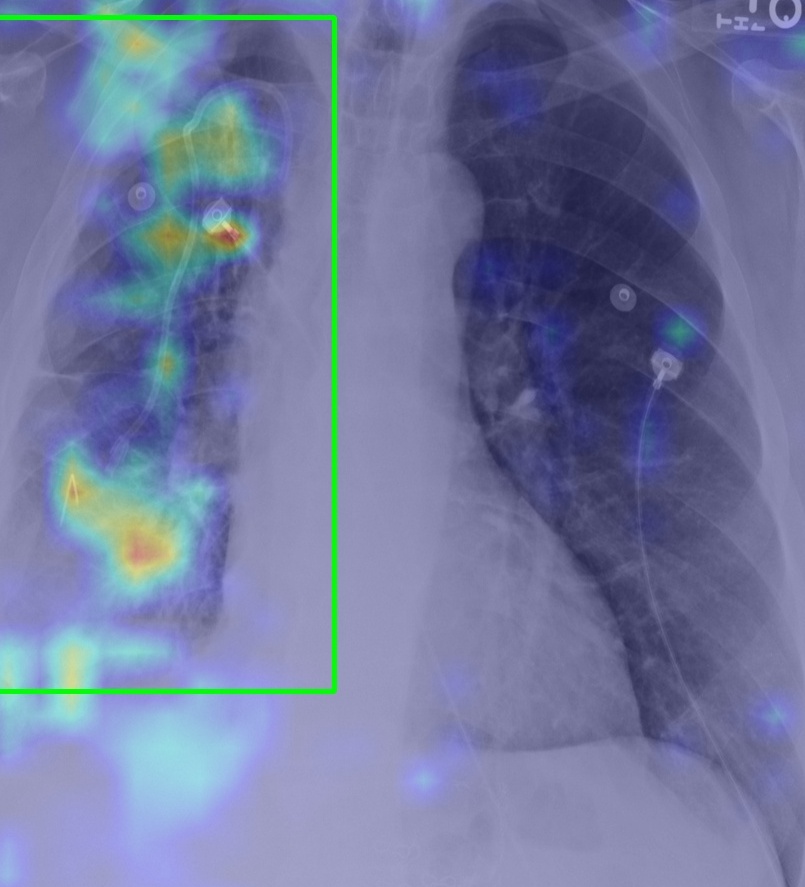}
        \label{fig:mass_cam}
    }


    \subfigure[Atelectasis, original]
    {
        \includegraphics[width=1.1in]{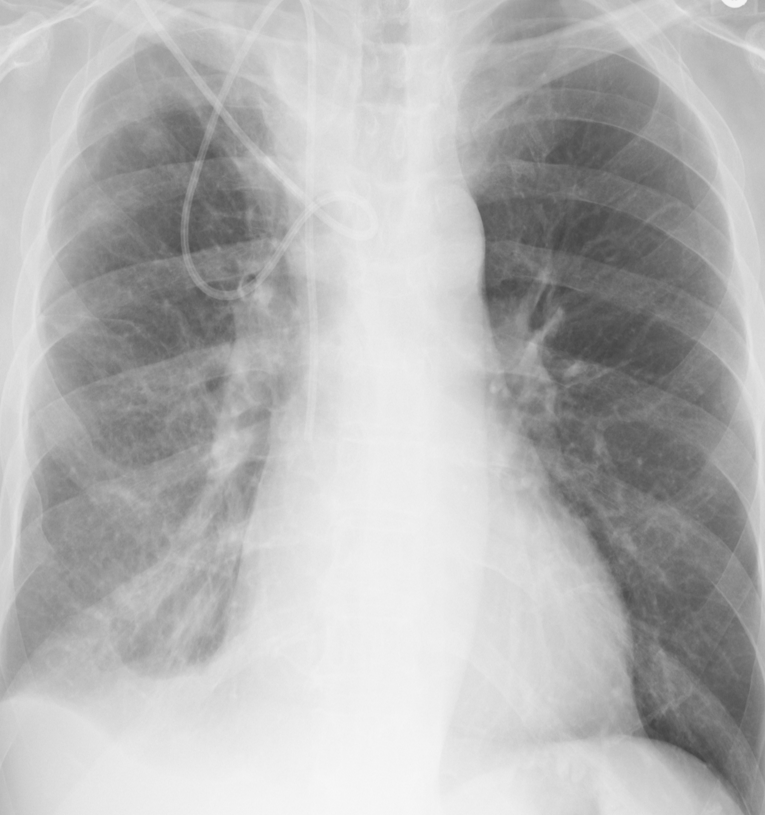}
        \label{fig:atelectasis1_orig}
    }
    \subfigure[Atelectasis, Grad-CAM + bbox]
    {
        \includegraphics[width=1.1in]{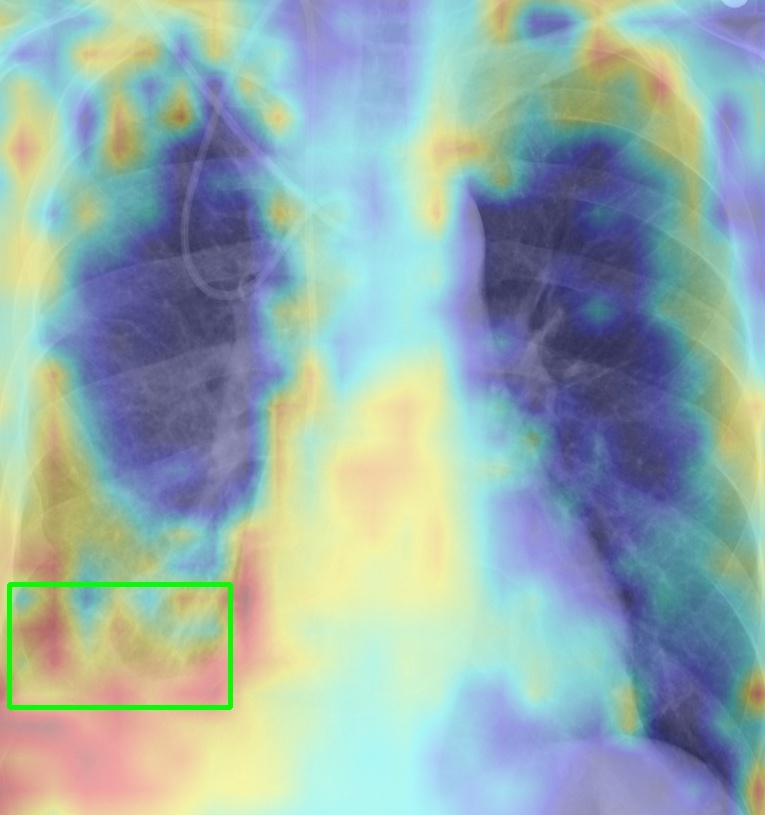}
        \label{fig:atelectasis1_cam}
    }


    \subfigure[Infiltration, original]
    {
        \includegraphics[width=1.1in]{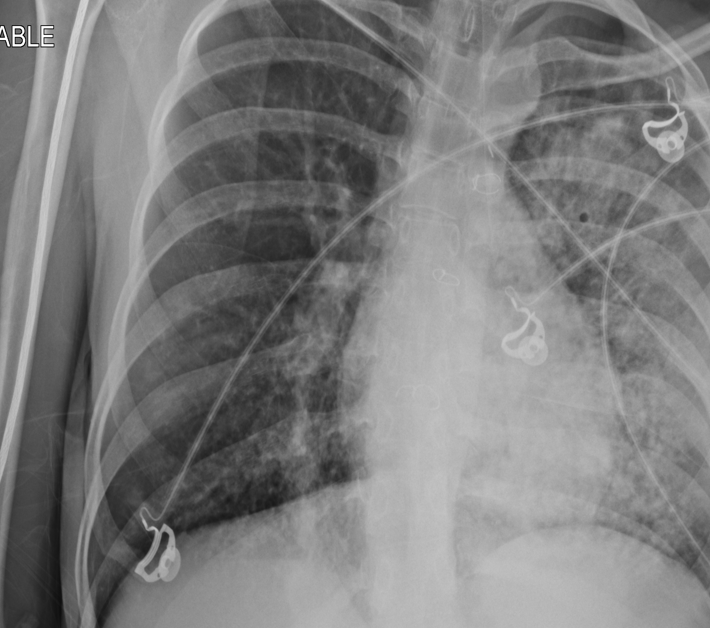}
        \label{fig:infiltrate1_orig}
    }
    \subfigure[Infiltration, Grad-CAM + bbox]
    {
        \includegraphics[width=1.1in]{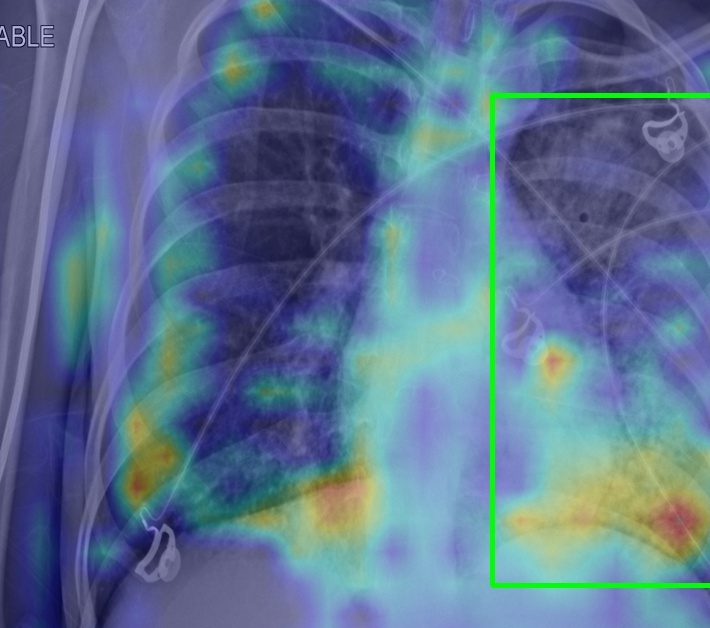}
        \label{fig:infiltrate1_cam}
    }


    \subfigure[Effusion, original]
    {
        \includegraphics[width=1.1in]{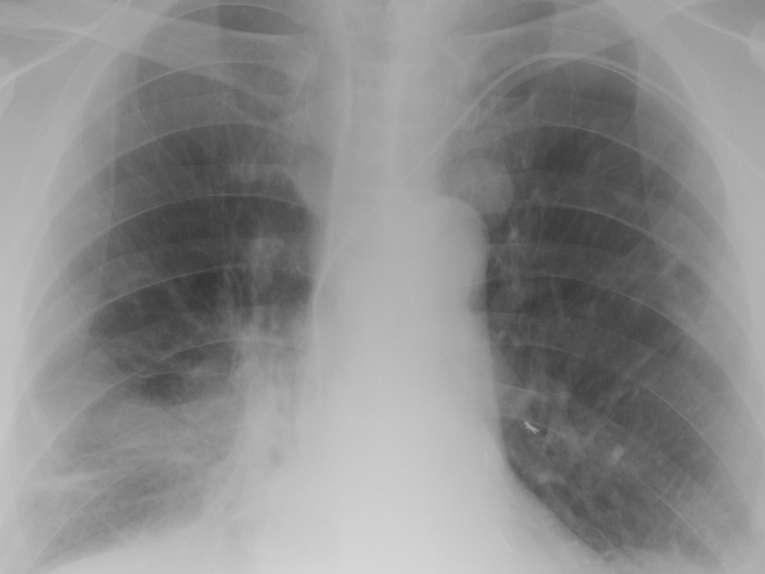}
        \label{fig:effusion_orig}
    }
    \subfigure[Effusion, Grad-CAM + bbox]
    {
        \includegraphics[width=1.1in]{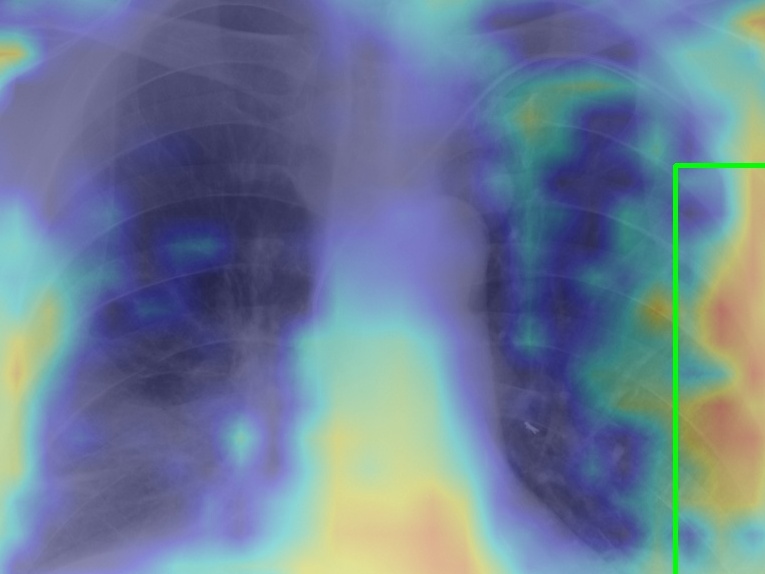}
        \label{fig:effusion_cam}
    }


    \subfigure[Pneumothorax, original]
    {
        \includegraphics[width=1.1in]{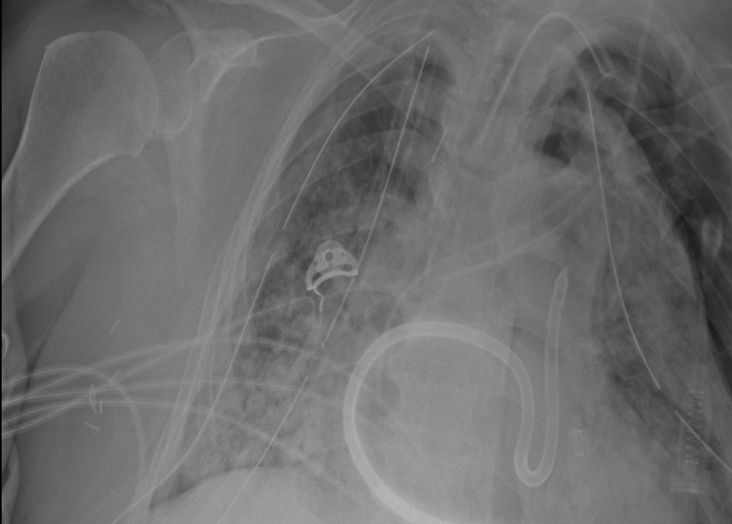}
        \label{fig:ptx_orig}
    }
    \subfigure[Pneumothorax, Grad-CAM + bbox]
    {
        \includegraphics[width=1.1in]{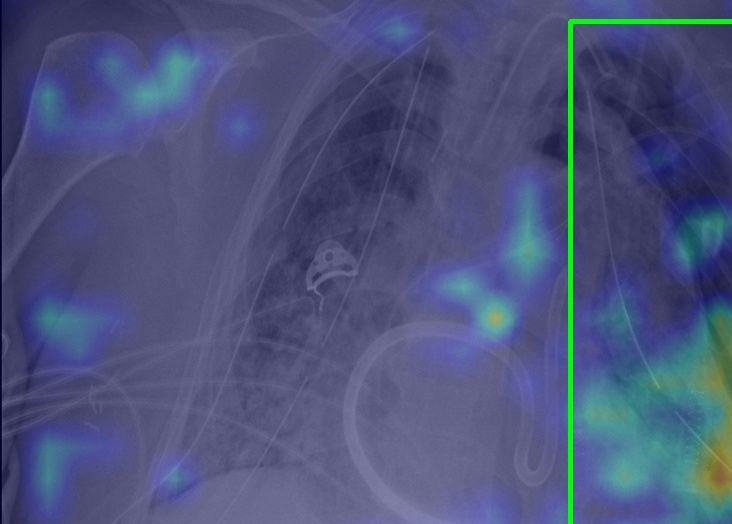}
        \label{fig:ptx_cam}
    }

    \caption{Symptom-level explainability examples. Each row represents a different symptom. The left image depicts the original chest X-ray, while the right image shows the Grad-CAM overlay highlighting areas of model focus. The green rectangle indicates the dataset-provided ground-truth bounding box for each symptom.}
    \label{fig:symptom_explain}
\end{figure}

To evaluate whether the model accurately identifies actual patterns in CXR images, we generated Grad-CAM \cite{gradcam2017} maps from the final transformer block for each positive symptom logit and overlaid them on the corresponding CXR images (Fig.~\ref{fig:symptom_explain}). In each image, the green rectangle represents the dataset-provided ground-truth bounding box for that symptom. We qualitatively assess the alignment of model saliency with these annotations, examining whether the highlighted areas correspond to the expected locations and appearance of each finding on chest radiographs.

\textbf{Fig.~\ref{fig:mass_cam}}. The heatmap reveals a compact, high-intensity focus within the annotated lung region, tapering off sharply at the edges. This pattern is consistent with a discrete rounded pulmonary opacity, characteristic of a mass, rather than diffuse parenchymal change. Saliency is primarily contained within the ground-truth bounding box.

\textbf{Fig.~\ref{fig:atelectasis1_cam}}.
The strongest saliency is observed at the lung base on the annotated side near the diaphragm, with activation overlapping the bounding box. The elongated, band-like basal pattern is typical of plate (discoid) atelectasis and indicative of regional volume loss.

\textbf{Fig.~\ref{fig:infiltrate1_cam}}.
The activation appears as a mottled, patchy distribution throughout the annotated lateral lung region, lacking a single round hotspot. This distribution could be characteristic of air-space disease, where alveolar filling leads to patchy opacities instead of a solitary nodule.

\textbf{Fig.~\ref{fig:effusion_cam}}.
Saliency is concentrated along the dependent lateral pleural region within the annotation, displaying a smooth, meniscus-like slope toward the costophrenic angle, which aligns with the expected contour of pleural fluid. Minor scattered signals above do not disrupt the dominant dependent pattern within the box.

\textbf{Fig.~\ref{fig:ptx_cam}}.
Saliency aligns with the apico-lateral pleural margin outlined by the narrow vertical annotation, while the adjacent, more lucent lung on the same side shows reduced activation. This corresponds to the expected location of the visceral pleural line on a pneumothorax film.

As shown in these five random examples, the model's attention concentrates on clinically meaningful regions that largely coincide with the dataset’s boxes. This qualitative concordance indicates that the symptom classification head leverages radiographically relevant evidence rather than spurious artifacts.

\section{Discussion and Conclusion}\label{sec:discuss}
This study introduces a DISTL-style teacher--student framework built on a ViT-Small backbone with two supervised heads for disease and symptom classification and a self-supervised DINO head. On the three-class disease classification task, the model achieves a high accuracy of 98.85\% and uniformly high class-wise F1 scores (normal 98.94\%, TB 99.44\%, COVID-19 98.30\%; see Table~\ref{tab:disease_cls}). For the multi-label symptom detection task, it excels all seven findings with a macro-F1 score of 90.09\% (Table~\ref{tab:sym}), outperforming four widely used CNN baselines. These advancements are attributed to three key design elements: (i) the joint optimization of disease and symptom objectives, which regularizes the backbone toward radiographic primitives; (ii) the DISTL schedule that mixes self-supervision with distillation before transitioning to pure distillation, improving sample efficiency under limited labels; and (iii) multi-crop training that exposes the network to both global lung context and fine-grained local views.

The symptom head proves particularly effective for detecting subtle or small targets, significantly boosting F1 scores for nodule (+40.6), effusion (+37.2), mass (+35.4), consolidation (+29.0), infiltration (+27.0), and atelectasis (+25.5), while also improving pneumothorax (+14.8) compared to the best CNN baseline. These improvements suggest that features learned for symptom recognition enhance disease classification, providing superior discrimination for TB/COVID than using disease labels alone. This supports workflows where disease triage is complemented by machine-generated radiographic evidence (e.g., “infiltration present” or “effusion present”), aiding in prioritizing readings and documenting findings.

Qualitative explainability analyses indicate that predictions are rooted in anatomically relevant areas. Grad-CAM maps, when compared to dataset bounding boxes (Fig.~\ref{fig:symptom_explain}), concentrate over focal intraparenchymal regions for ``mass'', basal band-like opacities for ``atelectasis'', patchy parenchymal changes for ``infiltration'', dependent lateral pleura for ``effusion'', and the apico-lateral pleural margin for ``pneumothorax''. Although some images exhibit minor off-box saliency (e.g., small superior hotspots), the predominant activation typically lies within or near the annotated region, supporting symptom-specific reasoning instead of non-diagnostic artifacts.

To conclude, the proposed single teacher--student ViT with joint disease and symptom supervision achieves state-of-the-art performance in TB/COVID/normal classification while simultaneously detecting diverse radiographic findings. Beyond aggregate metrics, the alignment between attention maps and annotated regions indicates that the model utilizes radiographically meaningful evidence. These properties render the approach a promising candidate for real-world screening and triage pipelines, where both reliable sensitivity for tuberculosis and actionable symptom readouts are crucial.

\section{Limitations and Future Work}\label{sec:limit}
This study has several limitations that warrant attention. First, the evaluation is retrospective and draws from multiple sources, which, despite patient-wise splits, may lead to distribution shifts due to label noise and inter-dataset heterogeneity in terms of projection, positioning, equipment, and acquisition protocols. Second, the symptom supervision relies on image-level labels and dataset-provided boxes that may be coarse; potentially resulting in partial localization without guaranteeing pixel-accurate delineation. Third, our analysis focuses on frontal CXRs, with no exploration of lateral views or computed tomography (CT) scans. Lastly, while the EMA teacher and pseudo-labeling help stabilize training, the model's performance might be sensitive to certain thresholds and the relative weighting between disease and symptom losses.

To address these limitations, future work will focus on several key areas: (i) conducting prospective, multi-site validation with site-specific calibration to assess the model's generalizability and robustness across different clinical settings; (ii) developing explicit localization heads trained with boxes or weak masks to complement Grad-CAM visualizations; (iii) implementing an adaptive curriculum for unlabeled data that utilizes confidence or uncertainty-aware sampling to improve model training efficiency; and (iv) integrating clinical metadata, such as age, symptoms, and prior imaging, to improve model calibration and robustness.

\begin{credits}
\subsubsection{\ackname} This study was funded by the New Beginnings program of the National Research Council of Canada.

\end{credits}
%
%
%
%

\end{document}